\documentclass[journal]{IEEEtran}
\pdfoutput=1
\usepackage{graphics}
\usepackage[]{graphicx}
\usepackage{amssymb,amsfonts,amsmath}
\usepackage{float}
\usepackage{cite}
\usepackage{balance}
\usepackage{threeparttable}
\usepackage{booktabs}
\usepackage{multirow}
\usepackage{subfigure}
\usepackage{tabularx,booktabs}
\usepackage{color}
\usepackage{amsmath}

\newcolumntype{L}{>{$}l<{$}}
\newcolumntype{C}{>{$}c<{$}}
\newcolumntype{R}{>{$}r<{$}}

\ifCLASSINFOpdf
\else
\fi
\begin{document}

\title{Movement Coordination in Human-Robot Teams: A Dynamical Systems Approach}
%\title{Synchronization Index Based Anticipatory Movement Selection in Human-Robot Teams}

\author{Tariq~Iqbal,
        Samantha Rack,
        and~Laurel~D.~Riek$^{1}$% <-this % stops a space 
{\tiny \thanks{The authors are from the Department of
Computer Science and Engineering, University of Notre Dame, Notre
Dame, IN, 46556 USA. 
e-mail: \{tiqbal, srack, lriek\}@nd.edu.}} 
}

\graphicspath{ {./Figures/} }
\pdfminorversion=4


\markboth{IEEE Transactions on Robotics,~Vol~32, Issue.~3, Jun~2016}%
{Iqbal, Rack, and Riek: Movement Coordination in Human-Robot Teams: A Dynamical Systems Approach}



\maketitle

%%%%%% Abstract
%%%%%%%%%%%%%%%%%%%%%%%%%%%%%%%%%%%%%%%%%%%%%%
\begin{abstract}
In order to be effective teammates, robots need to be able to understand high-level human behavior to recognize, anticipate, and adapt to human motion. 
We have designed a new approach to enable robots to perceive human group motion in real-time, anticipate future actions, and synthesize their own motion accordingly. 
We explore this within the context of \textit{joint action}, where humans and robots move together synchronously. 
In this paper, we present an anticipation method which takes high-level group behavior into account. 
We validate the method within a human-robot interaction scenario, where an autonomous mobile robot observes a team of human dancers, and then successfully and contingently coordinates its movements to ``join the dance''. We compared the results of our anticipation method to move the robot with another method which did not rely on high-level group behavior, and found our method performed better both in terms of more closely synchronizing the robot's motion to the team, and also exhibiting more contingent and fluent motion. These findings suggest that the robot performs better when it has an understanding of high-level group behavior than when it does not. This work will help enable others in the robotics community to build more fluent and adaptable robots in the future.

\end{abstract}

%\begin{IEEEkeywords}
%Human-Robot Interaction, Human Motion Analysis, Robot-based Motion Synthesis.
%\end{IEEEkeywords}

% \fi
\IEEEpeerreviewmaketitle

%%%%%%%%%%%%%%%%%%%%%%%%%%%%%%%%%%%%%%%%%%%%%%
%%%%%% Section: 1
%%%%%% Introduction
%%%%%%%%%%%%%%%%%%%%%%%%%%%%%%%%%%%%%%%%%%%%%%

\section{Introduction}
As technology advances, autonomous robots are becoming more involved in human society in a variety of roles.
Robotic systems have long been involved in assembly lines automating and increasing efficiency of monotonous or dangerous factory procedures \cite{Wilcox2012}.
However, as robots leave controlled spaces and begin to work alongside people in teams, many things taken for granted in robotics concerning perception and action do not apply, as people act unpredictably, e.g., they ``break the rules" when it comes to what a robot can expect \textit{a priori}.
In order for robots to effectively perform their tasks and integrate in Human Social Environments (HSEs), they must be able to comprehend high-level social signals and respond appropriately \cite{Riek2013}.

While working alongside humans, a robot might encounter people performing various social actions, such as engaging in social activities, or performing synchronous movements \cite{Rack2015HRI}.
For example, Ros et al. \cite{Ros2014} used a humanoid robot to play the role of a dance instructor with children, and Fasola et al. \cite{Fasola2013} designed a socially assistive robot to engage older adults in physical exercise. Others have used robots to dance and play cooperatively with children in therapeutic  settings \cite{Michalowski2007, Park2010}. 
Koenemann et al. demonstrated a system which enabled humanoid robots to imitate complex human whole-body motion \cite{Koenemann2014}.

However, sometimes it can be difficult for a robot to perceive and understand all of the different types of events involved during these activities to make effective decisions, due to sensor occlusion, unanticipated motion, narrow field of view, etc.
On the other hand, if a robot is able to make better sense of its environment and understand high-level group dynamics, then it can make effective decisions about its actions.
If the robot has this understanding of its environment, then its interactions within the team might reach to a higher-level of coordination, resulting in a \textit{fluent} meshing of actions \cite{Riek2010b, Iqbal2015ROMAN, Hoffman2007TRO, cakmak2011using}. 

Human activity recognition from body movement is an active area of research across many fields \cite{Bandouch2012,Zhang2011,Koppula2013,Ofli2013,Samadani2013}.
These activities involve a wide range of behaviors, from gross motor motion (e.g., walking, lifting) to manipulation (e.g., stacking objects).
All of these experiments showed impressive results in recognizing activities, either performed by individual or dyad.

However, the focus of most of these methods are to recognize the activity of a single human, rather than to understand the a team's dynamics and how it might affect behavior. This understanding is critical in human-robot interaction scenarios, as the ``one human, one robot'' paradigm is rarely seen in ecological settings \cite{burgard1999experiences,aastha}. To make informed decisions, robots need to understand this context \cite{Iqbal2015ROMAN}.

%{\color{red}{
Many disciplines have investigated interaction dynamics within groups, which include sociology, psychology, biology, music and dance \cite{Nagy2010,Nagy2013,Leonard2012,Leonard2014, Clayton2005,Schmidt2008,Pereda2005,Khalfa2008, Kreuz2007,Himberg2011,Lakens2010,Hennig2014}. For example, Nagy et al. \cite{Nagy2010, Nagy2013} investigated collective behavior on animals, and developed automated methods for assessing social dominance and leadership in domestic pigeons. Their investigation explored the effect of social hierarchical structure on dominance and leadership. Their results indicated that dominance and leadership hierarchical structures were independent from each other. 

Inspired from bird flocks and fish schools, Leonard et al. \cite{Leonard2012, Leonard2014}  investigated how collective group motion emerges when basic animal flocking rules (i.e., cohesive and repulsive element) are applied on a group of human dancers. Using tracked trajectories of head positions of individual dancers, the authors developed a time-varying graph-based method to infer conditions under which certain dancers emerged as the leaders of the group.

%}}

Synchronous motion, or joint action, is a common type of high-level behavior encountered in human-human interaction.
It is a naturally present social interaction that occurs when two or more participants coordinate their actions in space and time to make changes to the environment \cite{Sebanz2006}. 
Understanding synchronous joint action is important, as it helps to accurately understand the affective behavior of a team, and also provides information regarding the group level cohesiveness \cite{Richardson2012,Varni2010}.
Thus, if a robot has the ability to understand the presence of synchronous joint action in a team, then it can use that information to inform its own actions to enable coordinated movement with the team. It also might learn advanced adaptive coordination techniques the human teams use, such as tempo adaptation or cross-training \cite{VanDerSteen2015,Wilcox2012}. 
%For example, this understanding will help the robot to perform the task synchronously with the rest of the group members.
%Moreover, this understanding might lead to a robot to achieve the ability of synchronizing its movements with team activities containing tempo changes, which is a natural skill exists in humans \cite{VanDerSteen2015}.

%{\color{blue}{
Many approaches have been taken by  researchers across different fields to measure the degree of synchronization in continuous time series data, including recurrence analysis \cite{Marwan2007}, correlation \cite{Konvalinka2010}, and phase difference approaches \cite{Richardson2012}. Other sets of methods work across categorical time series data, which may define discrete events \cite{QuianQuiroga2002}. However, these event-based methods only consider a single type of event while measuring synchronization. To address this gap, we created an event-based method which can successfully take multiple types of discrete, task-level events into consideration while measuring the degree of synchronization of a system \cite{Iqbal2015TAC}. 
%}}

%Based on its understanding on the environment, if a robot can make predictions about the future actions of its human counterparts, then the robot is capable of performing more fluent interaction \cite{Hoffman2007HRI}.
Recent work in robotics has focused on developing predictive methods for improving the fluency of a joint interaction between a robot and one or more humans.
For example, Hawkins et. al.  \cite{Hawkins2014} developed a method that determines an appropriate action for an assistive robot to take when providing parts during an assembly activity. They employ a probabilistic model that considers the presence of variability in the human's actions.
Hoffman et al. \cite{Hoffman2007TRO} proposed an adaptive action selection mechanism for a robot, which could make anticipatory decisions based on confidence of their validity and their relative risks. Through a study, the authors validated the model and presented an improvement in task efficiency when compared to a purely reactive model. 

Additionally, P{\'e}rez-D'Arpino et al. \cite{Perez2015ICRA} proposed a data-driven approach to synthesize anticipatory knowledge of human motion, which they used to predict targets during reaching motions.
Unhelkar et al. \cite{Unhelkar2015ICRA} extended this concept for a human-robot co-navigation task.
This model used an ``human turn signals'' during walking as anticipatory indicators, in order to predict human motion trajectories. This knowledge was then used for motion planning in simulated dynamic environments.

%%%%%%%%%%%%%%%%%%%%%%%%%%%%%%%%%%%%%%%%%%%%%%
%%%%%% Figure: Setup
%%%%%%%%%%%%%%%%%%%%%%%%%%%%%%%%%%%%%%%%%%%%%%
\begin{figure}[t]
\centerline{\includegraphics[width=.45\textwidth]{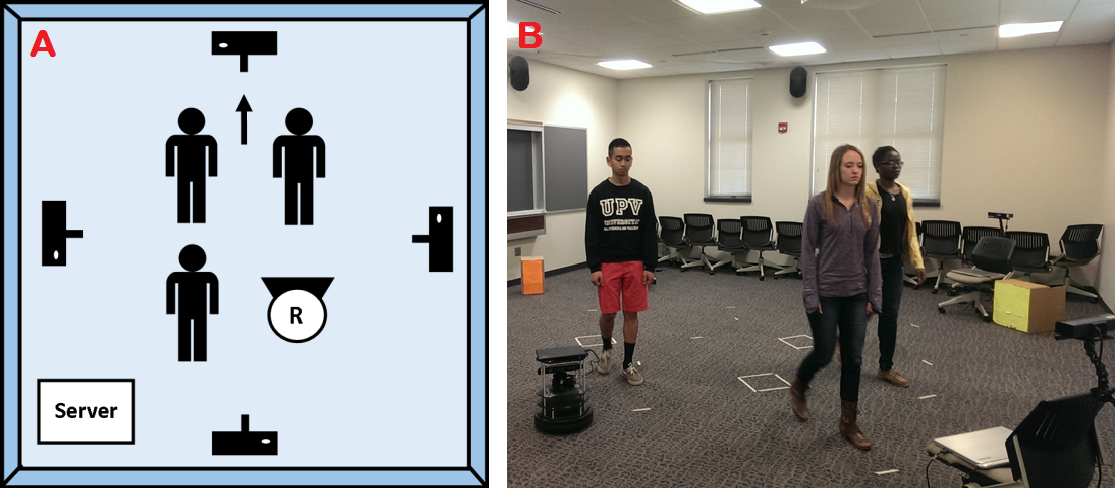}}
\caption{A) Data acquisition setup. B) Three participants are dancing along with a Turtlebot robot.}
\label{fig:setup}
\vspace{-0.15in}
\end{figure}

While this work will improve the ability of robots to have fluent interactions within HSEs, most of these methods are best-suited for dyadic interaction and dexterous manipulation contexts. In contrast, we seek to explore methods for robots that will work robustly in groups, and also for tasks involving gross motion with mobile robots. 
%We also seek to explore methods that will work in real time, in settings as ecologically realistic as possible.

%Thus, we aim to build a method for a robot which will take the high-level understanding of the group behavior, be able to anticipate future activities of that group based on this understanding, and perform accordingly.

In our prior work, we explored the problem of automatically modelling and detecting synchronous joint action (SJA) in human teams, using both fixed and mobile sensors. We introduced a new, non-linear dynamical method which performed more accurately and robustly than existing methods \cite{Iqbal2015TAC,Iqbal2015ROMAN}.

In this paper, we explore how a robot can use these models to synthesize SJA in order to coordinate its movements with a human team. 
%We present an SJA-based anticipation method which takes high-level group behavior into account. 
The main contribution of this work is the introduction of a new method for anticipation of robot motion that takes human group dynamics into account.
We validated our method within a human-robot interaction scenario, where an autonomous mobile robot observes a team of human dancers, and then successfully and contingently coordinates its movements to ``join the dance''. We compared the results of our anticipation method with another method that does not rely on high-level group behavior. Our method performed better both in terms of more closely synchronizing the robot's motion to the team, and exhibiting more contingent and fluent motion.

The outline of the paper is as follows. Section~\ref{sec:setup}, describes the experimental testbed for studying SJA and the system architecture. Then, Section~\ref{sec:method} provides details of the two anticipation methods. Section~\ref{sec:experiment} describes the experimental procedure. 
Sections~\ref{sec:analysis} and ~\ref{sec:result} discuss how the data were pre-proessed, and the experimental results. Finally, Section~\ref{sec:discussion} discusses the implication of these findings for the robotics community.

%%%%%%%%%%%%%%%%%%%%%%%%%%%%%%%%%%%%%%%%%%%%%%
%%%%%% Section: 2
%%%%%% System Architecture and Experimental Testbed
%%%%%%%%%%%%%%%%%%%%%%%%%%%%%%%%%%%%%%%%%%%%%%
%\section{System Architecture and Methodology}\label{sec:setup}
\vspace{-0.05in}
\section{System Architecture and Experimental Testbed}\label{sec:setup}

In order to explore how a robot can use  human group dynamics to synthesize SJA with a mixed team, we needed an experimental testbed where a robot could perform tasks synchronously with humans. We also required a group activity where each member's actions would have impact on others' actions, as well as have impact on the dynamics of the group overall.

\begin{figure*}[t]
\centerline{\includegraphics[width=0.90\textwidth]{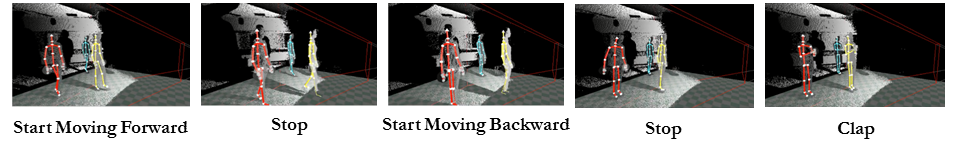}}
\caption{Five high-level were detected during the dance using skeletal data from participants. One iteration of the dance consists of two passes of the \textit{start moving forward}, \textit{stop moving forward}, \textit{start moving backward}, \textit{stop moving backward} events, and a \textit{clap} event in order.}
\label{fig:events}
\vspace{-0.15in}
\end{figure*}

Therefore, we designed a movement task where a team of humans and a robot could coordinate their motion in real-time. Specifically, we explored SJA within the context of synchronous dance. In concert with an experienced dancer, we choreographed a routine to the song \textit{Smooth Criminal} by Michael Jackson, which is in 4/4 time. The dance is iterative, and performed cyclically in a counter-clockwise manner (see Figure~\ref{fig:setup}-A.) There are four \textit{iterations} in a dance session, corresponding to each of the cardinal directions (North, West, South, and East). Each iteration includes the dancers taking the following steps in order: move forward and backward twice, then, clap, and turn 90-degrees (see Figure~\ref{fig:events}) \cite{Rack2015HRI}.

%%%%%%%%%%%%%%%%%%%%%%%%%%%%%%%%%%%%%%%%%%%%%%
%%%%%% Sub Section: 2.1 
%%%%%% Data Acquisition Process
%%%%%%%%%%%%%%%%%%%%%%%%%%%%%%%%%%%%%%%%%%%%%%

%%%%%%%%%%%%%%%%%%%%%%%%%%%%%%%%%%%%%%%%%%%%%%
%%%%%% Figure: Events
%%%%%%%%%%%%%%%%%%%%%%%%%%%%%%%%%%%%%%%%%%%%%%
%\begin{figure*}[t]
%\centerline{\includegraphics[width=.85\textwidth]{events1.png}}
%\caption{High-level events during one iteration of the dance movements over time. One iteration of the dance session consists of move forward, move backward, move forward, move backward, clap, and a 90-degree turn in order.}
%\label{fig:events}
%\hspace{-0.15in}
%\end{figure*}

%%%%%%%%%%%%%%%%%%%%%%%%%%%%%%%%%%%%%%%%%%%%%%
%%%%%% End Figure: Events
%%%%%%%%%%%%%%%%%%%%%%%%%%%%%%%%%%%%%%%%%%%%%%

\subsection{Data Acquisition Process}
Figure~\ref{fig:setup}-A shows the data acquisition setup. Three human participants and a Turtlebot v.2 robot were arranged in two rows. Four Microsoft Kinect v.2 sensors were positioned approximately three feet above the ground at each of the cardinal directions. Each sensor was connected to a computer (client) to capture and process the depth, infrared, and skeletal data from the Kinect. All four clients and the server ran Windows 8 on an Intel Core i5 processor at 1.70Hz with 12GB of RAM.

As we are studying synchronous activity, it was critical all clients and the robot maintained a consistent time reference. Thus, we created a server to manage communication and global time synchronization.  Synchronization architecture details can be found in Iqbal et al. \cite{Iqbal2014HRI}.  

Each client performed real-time processing of the raw data in order to detect dance events (e.g., move forward, stop, etc), which it sent to the server with a timestamp. When the server received data from the clients, it generated predictions for how the robot should move using one of two anticipation methods, which are described in Section \ref{sec:method}. The server was also responsible for determining the \textit{active client}, which refers to which of the four sensors the dancers were facing during a given iteration. 

%The robot node runs on the Robot Operating System (ROS) platform to control a Turtlebot's movements based on instructions received from the server.

In order to allow for offline analysis, the clients also recorded time-synchronized depth, infrared, audio, and skeletal data using at automated interface with Kinect Studio. The server and robot also kept detailed logs of all communication, odometry information, events received from the active client, and information about the dancers.

%%%%%%%%%%%%%%%%%%%%%%%%%%%%%%%%%%%%%%%%%%%%%%
%%%%%% Sub Section: 2.2
%%%%%% Data Processing in Clients
%%%%%%%%%%%%%%%%%%%%%%%%%%%%%%%%%%%%%%%%%%%%%%
\subsection{Client-side data processing}
%Each client processes the data acquired from its connected Kinect sensor.
%From the processing of the participants' body movements extracted from the data streams, the clients detect the high-level events involved in the dance.
%When an event is detected, the clients send event data with the time at which it occurred and other information (e.g., velocity) to the server.
%Based on the data sent from the clients in earlier stages of the dance, the server predicts the future movements for the robot.
%In the next Section, we will discuss the event detection method from the client end. 

%%%%%%%%%%%%%%%%%%%%%%%%%%%%%%%%%%%%%%%%%%%%%%
%%%%%% High-level Event Detection
%%%%%%%%%%%%%%%%%%%%%%%%%%%%%%%%%%%%%%%%%%%%%%
%\subsubsection{High-level event detection} \label{sec:event_detection}
We extracted five high-level events from the participants' movements during the dance: \textit{start moving forward}, \textit{stop moving forward}, \textit{start moving backward}, \textit{stop moving backward}, and \textit{clap}. The \textit{start moving forward} event is detected when a participant begins approaching the Kinect, and \textit{stop moving forward} when they stop moving. Similarly, as a participant moves away from the sensor (backward), that is identified as a \textit{start moving backward} event, and when they stop, \textit{stop moving backward}. We also detected participants' \textit{clap} events, which occurred at the end of each iteration. See Figure~\ref{fig:events}.

To detect these events from participants' body movements, clients used the skeletal positions provided by the Kinect. Clients calculated forward and backward motion onsets along the \textit{z-axis} primarily using the spine base body joint position, as it is the most stable and reliable joint position when participants are in motion. 

However, there were times when participants did not move their spine base, but did move their mid-spine, shoulders, or neck, to signal the onset of motion. Therefore, clients also used these positions, again along the \textit{z-axis}, to detect four additional events: \textit{early start moving forward}, \textit{early stop moving forward}, \textit{early start moving backward}, and \textit{early stop moving backward}. For these early events, clients calculated joint change positions by comparing the current and previous frame. If at least half of the joint positions changed, then it indicated the participant had started moving. To detect  \textit{clap} events, clients used the x and y coordinates from the 3D skeletal position of the left and right hand and shoulder joints. Claps occurred when the ratio of the distance between the hands and the distance between the shoulder joints was less than a threshold (0.6), and when this ratio value reaches a local minima.

%%%%%%%%%%%%%%%%%%%%%%%%%%%%%%%%%%%%%%%%%%%%%%
%%%%%% Figure: Methods
%%%%%%%%%%%%%%%%%%%%%%%%%%%%%%%%%%%%%%%%%%%%%%

\begin{figure*}[t]
\centerline{\includegraphics[width=0.8\textwidth]{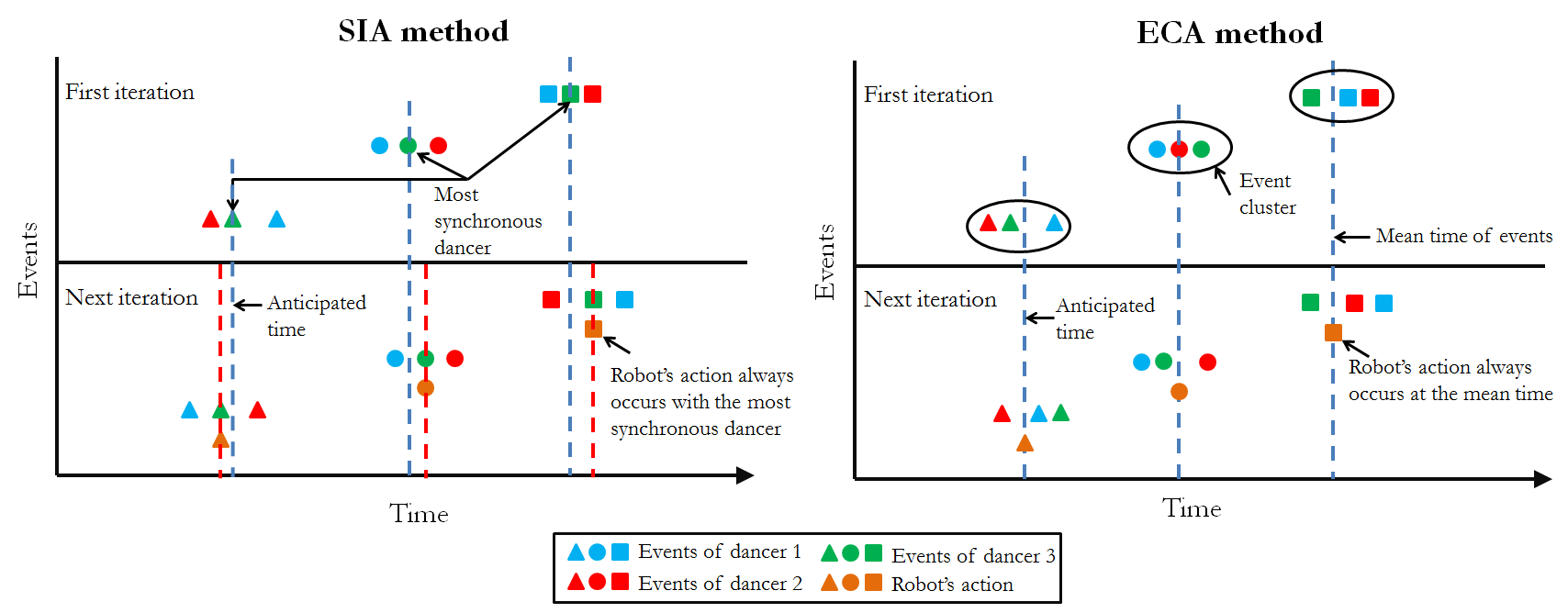}}
\caption{A visualization of the two anticipation methods. Left: Synchronization Index Based Anticipation (SIA), Right: Event Cluster Based Anticipation (ECA). The upper portion of the graph represents one iteration of the dance session, and the lower portion represents the next iteration of the same session.}
\label{fig:methods}
\vspace{-0.15in}
\end{figure*}

%%%%%%%%%%%%%%%%%%%%%%%%%%%%%%%%%%%%%%%%%%%%%%
%%%%%% Sub Section: 2.4
%%%%%% Robot Command Generation
%%%%%%%%%%%%%%%%%%%%%%%%%%%%%%%%%%%%%%%%%%%%%%
\subsection{Robot Command Generation and Execution}
After the server determines which movement the robot should make, which it does using an anticipation methods described in \ref{sec:method}, it sends a movement command to the robot. These commands include: \textit{move forward}, \textit{move backward}, \textit{stop}, and \textit{turn}. The server translated the clap commands into rotation commands while sending it to the robot, since the robot can't clap. 

%After predicting the future events, or receiving commands with real time events for the robot, the server sends these movement commands at the appropriate time to the robot.
%If needed, the server also sends additional data calculated from the previously sent events to the robot, e.g., velocity for forward or backward movement.

The robot, which ran the Robot Operating System (ROS) version Hydro on Ubuntu version 12.04, accepted commands from the server, parsed the commands, and used an ROS publisher to send movement commands to the controller. The robot is capable of forward and backward movement, and can rotate on its vertical axis in either direction.

%%%%%%%%%%%%%%%%%%%%%%%%%%%%%%%%%%%%%%%%%%%%%%
%%%%%% Section: 3
%%%%%% Event Prediction Methods
%%%%%%%%%%%%%%%%%%%%%%%%%%%%%%%%%%%%%%%%%%%%%%
\section{Event Anticipation Methods} \label{sec:method}

For this work, we created two anticipation methods to move the robot. The first method, synchronization-index based anticipation (SIA), is inspired by our prior SJA detection work \cite{Iqbal2015TAC}. It calculates the synchronicity of the group in real-time, determines who the most synchronous dancer is, and uses that information to move the robot. The second method, event cluster based anticipation (ECA), we created to establish a reasonable comparison anticipation method for SIA that does not rely on group dynamics. ECA is a straightforward method that involves averaging the times participants moved during a previous iteration of the dance. Figure~\ref{fig:methods} gives a visual comparison of how the two methods work in practice, and they are described textually below.

%%%%%%%%%%%%%%%%%%%%%%%%%%%%%%%%%%%%%%%%%%%%%%
%%%%%% Sub Section: 3.1
%%%%%% SIA
%%%%%%%%%%%%%%%%%%%%%%%%%%%%%%%%%%%%%%%%%%%%%%
\vspace{-0.06in}
\subsection{Synchronization Index Based Anticipation (SIA)}\label{sec:sia}
The SIA method takes a group's internal dynamics into account when generating robot movements. The main idea is that for a given iteration, the participant who moves the most synchronously with the other dancers is a good model for the robot to follow in order to be well-coordinated with the team. %Furthermore, the method will adjust after each iteration, as who the most synchronous dancer can change over time. 
Also, the method will adjust its identification of the most synchronous dancer after each iteration.
Figure~\ref{fig:methods}-B explains this method visually.

Thus, to generate future actions for the robot using this method, at the beginning of each iteration we measured the most synchronous person of the group using our the non-linear dynamical method we described in Iqbal and Riek \cite{Iqbal2015TAC}. We will briefly describe the method in Sections~\ref{sec:measuringSync} and \ref{indiv}, and then discuss in Section~\ref{anticipate} how we used the method to assess the most synchronous dancer to inform how the robot should move. 
%First, we will describe the method to measure synchronization of a pair of people for single types of events, then extend the method for multiple types of events. 
%Then, we will describe the method to measure individual and group synchronization index.

%%%%%%%%%%%%%%%%%%%%%%%%%%%%%%%%%%%%%%%%%%%%%%
%%%%%% Sub Section: 3.1.1
%%%%%%%%%%%%%%%%%%%%%%%%%%%%%%%%%%%%%%%%%%%%%%
\subsubsection{Measuring synchronization of events across two time series} \label{sec:measuringSync}
We can express the task-level events associated with each dancer as a time series. Suppose ${x_n}$ and ${y_n}$ are two time series, where $n=1 \ldots N$. Here, each time series has $N$ samples. Suppose, $m_x$ and $m_y$ are the number of events occuring in time series $x$ and $y$ respectively, and $E$ is the set of all events \cite{Iqbal2015TAC}.

The events of both series are denoted by $e_x(i) \in E$ and $e_y(j) \in E$,  where, $i=1 \ldots m_x$, $j=1 \ldots m_y$. The event times on both time series are $t_i^x$ and $t_j^y$ ($i = 1 \ldots m_x$, $j = 1 \ldots m_y$) respectively \cite{Iqbal2015TAC}.

In the case of synchronous events in both time series, the same event should appear roughly at the same time, or within a time lag $\pm\tau$ \cite{Iqbal2015TAC}.

Now, suppose $c^\tau(x|y)$ denotes the number of times a single type of event $e \in E$ appear in time series $x$ shortly after they appear in time series $y$. Here,

\vspace{-0.08in}
\begin{equation}
c^\tau(x|y)  = \sum_i^{m_x}\sum_j^{m_y}J_{ij}^\tau
\end{equation}
where,
\begin{equation}
J_{ij}^\tau = \left\{ \begin{array}{ll}
1 & \mbox{if $0 < t_i^x - t_j^y < \tau$}\\
\frac{1}{2} & \mbox{if $t_i^x = t_j^y$}\\
0 & \mbox{otherwise} \end{array} \right. 
\end{equation}
\vspace{-0.08in}

Similarly, we can calculate $c^\tau(y|x)$ denoting the number of times a single type of event $e \in E$ appear in time series $y$ shortly after they appear in time series $x$. 

Now, $Q_\tau(e)$ represents the synchronization of events in two time series, where we are only considering a single type of event $e$ in both time series. From $c^\tau(x|y)$  and $c^\tau(y|x)$, we can calculate $Q_\tau(e)$ as,

\begin{equation}
Q_\tau (e)  = \frac{c^\tau(x|y)+c^\tau(y|x)}{\sqrt{m_x m_y}}
\end{equation}

The value of $Q_\tau(e)$ should be in between $0$ and $1$ ($0~\leq~Q_\tau(e)~\leq~1$), as we normalize it by the number of events that appear in both time series.
$Q_\tau(e)=1$ shows that all the events of both time series are fully synchronized, and appeared within a time lag $\pm\tau$ on both time series. On the other hand, $Q_\tau(e)=0$ shows us that the events are asynchronous \cite{Iqbal2015TAC}.

Now, we extend the notion of synchronization of events in two time series for multiple types of events. 
Suppose we have $n$ types of events $\{e_1, e_2,\ldots,e_n\} \in E(n)$, where $E(n)$ is the set of all types of events.
First, we calculate $Q_\tau(e_i)$ for each event type $e_i \in E(n)$. While calculating $Q_\tau(e_i)$, we will not consider any other type of event, except $e_i$ \cite{Iqbal2015TAC}.

Now, let $m_x(e_i)$ be the number of events of type $e_i$ occurring in time series $x$ and $m_y(e_i)$ is the number of events of type $e_i$ occurring in time series $y$.
To measure synchronization of multiple types of events between two time series, we take the average of $Q_\tau(e_i)$ for each event type $e_i$, weighted by the number of events of that type. We will call this the synchronization index of that pair \cite{Iqbal2015TAC}.

So, the overall synchronization of events in time series $x$ and $y$ of that pair is:

\vspace{-0.08in}
\begin{equation}\label{eq:1}
\forall  e_i \in E(n) : Q_\tau^{xy} = \frac{\sum{[Q_\tau(e_i)  \times [m_x(e_i)+m_y(e_i)]]}}{\sum{[m_x(e_i)+m_y(e_i)]}} 
\end{equation}
\vspace{-0.08in}

If all events are synchronous in both time series, then the value of $Q_\tau^{xy}$ will be 1. If no synchronous are synchronous, the value of $Q_\tau^{xy}$ will be 0 \cite{Iqbal2015TAC}.

%%%%%%%%%%%%%%%%%%%%%%%%%%%%%%%%%%%%%%%%%%%%%%
%%%%%% Sub Section: 3.1.2
%%%%%% Measuring indiv. sync
%%%%%%%%%%%%%%%%%%%%%%%%%%%%%%%%%%%%%%%%%%%%%%
\subsubsection{Measuring the individual synchronization index}
\label{indiv}

We calculated the pairwise synchronization index for each pair. Suppose we have $H$ number of time series. 
The time series data are represented as $s_1, s_2,\ldots,s_H$. First, we calculate the pairwise event synchronization index for each pair. So, we have the value of $Q_\tau^{s_{1}s_{2}}, Q_\tau^{s_{1}s_{3}},\ldots,Q_\tau^{s_{(H-1)}s_{H}}$ \cite{Iqbal2015TAC}. 

We modified our process slightly from the description in Iqbal and Riek \cite{Iqbal2015TAC}. After calculating the pairwise synchronization index, we built a directed weighted graph from these indices, where each time series is represented by a vertex.
However, in \cite{Iqbal2015TAC}, after calculating the pairwise synchronization index, an undirected weighted graph was built.
In a fully connected situation, the directed and the undirected graph represents the same connectivity.

So, if the time series are $s_1, s_2,\ldots,s_H$, then there is a vertex in the graph which will correspond to a time series. We connect a pair of vertices with a weighted edge, based on their synchronization index value.
In this case, there will be an incoming and an outgoing edge for each pair of vertices.
We will refer to this graph as the \textit{group topology graph (GTG)} \cite{Iqbal2015TAC}.

The individual synchronization index ($I_\tau(s_i)$) depends on both the group composition and the size of the group. We assumed that during this dance performance, each human participant may have some direct or indirect influences on the other human participants of the group \cite{Iqbal2015TAC}. $I_\tau(s_i)$ for a participant is measured as the average of the weight of the outgoing edges to the corresponding vertex in the topology graph. So, the $I_\tau(s_i)$ of series $s_i$ is:

\vspace{-0.08in}
\begin{equation}
%\[
I_\tau(s_i)  = \frac{\sum_{j=1,\ldots,H, \; j \neq i}{Q_\tau^{s_{{i}}s_{{j}}}} \times f(s_i,s_j)}{\sum_{j=1,\ldots,H, \; j \neq i} f(s_i,s_j)}
%\]
\end{equation}

Where,
\begin{equation}
%\[
f(s_i,s_j) = \left\{ \begin{array}{ll}
         1 & \mbox{iff \textit{edge}$(s_i,s_j)$ $\in$ $GTG$}\\
         0 & \mbox{otherwise} \end{array} \right. 
%\]
\end{equation}
\vspace{-0.08in}

%%%%%%%%%%%%%%%%%%%%%%%%%%%%%%%%%%%%%%%%%%%%%%
%%%%%% Sub Section: 3.1.3
%%%%%%%%%%%%%%%%%%%%%%%%%%%%%%%%%%%%%%%%%%%%%%
\subsubsection{Determining the most synchronous dancer and anticipating their next movement}
\label{anticipate}

The person with the highest individual synchronization index during an iteration is considered the most synchronous person of the group. This is because a high individual synchronization index indicates close synchronization with the other group members. Thus, let this person be $MSP$.

%%%%%%%%% Justification of following the $MSP$ person

%%%%%%%%%%%%%%%%%%%%%%%%%%%%%%%%%%%%%%%%%%%%%%%%%%
%%%%%%%%%%%%%%%%%%%%%%%%%%%%%%%%%%%%%%%%%%%%%%%%%%

Suppose, during $itr_i$, we determine $MSP(itr_i)$ as the most synchronous dancer of the group. Now, assuming that a similar timing pattern of events will occur during the next iteration ($itr_{(i+1)}$), if the robot follows the events of the $MSP(itr_i)$, then the group will become more synchronous.
%then the group will reach to the most synchronous state.

We can describe this concept mathematically. 
%To reach a synchronous state, all events must occur very closely in time. 
To reach a synchronous state, all events must occur very closely in time, i.e., within a time lag $\pm\tau$. Thus, we want to minimize the difference between event timings for each pair of agents. Now, if $\Delta t_{ij}$ represents the time difference of one event between agent $i$ and $j$, then our goal is:
\begin{equation}
\vspace{-\topsep}
\forall  i, j \in H : min \: (\sum{\Delta t_{ij}})
\end{equation}

\begin{figure}[t]
\centerline{\includegraphics[width=0.40\textwidth]{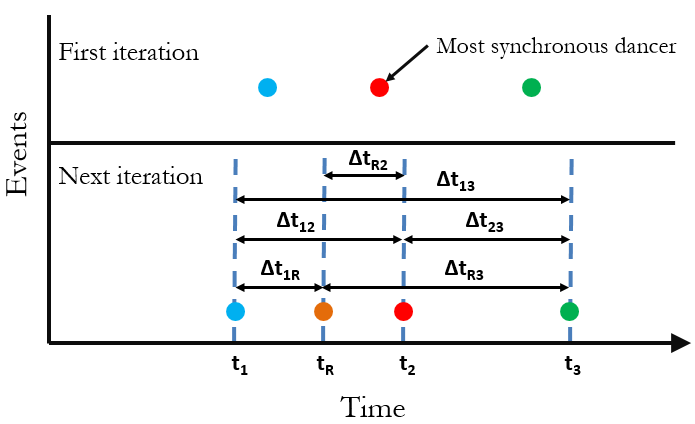}}
\caption{Example timings from a single type of event during two consecutive iterations.}
\label{fig:justificationSIA}
\vspace{-0.15in}
\end{figure}

Now for our scenario, as shown in Figure~\ref{fig:justificationSIA},  suppose $Dancer \: 2$ was the most synchronous person during one iteration ($itr_i$) of the dance session, i.e., $MSP(itr_i)$ was $Dancer \: 2$. Now, during $itr_{(i+1)}$, a similar timing pattern holds, and the timing of one particular event of the three dancers and the robot are $t_1$, $t_2$, $t_3$, and $t_R$ respectively. To reach a synchronous state, the following is required:
\vspace{-0.08in}
\begin{equation} \label{eq:justificationSIA}
\begin{split}
min \:  & (\sum{\Delta t_{12} + \Delta t_{23} + \Delta t_{1R}}  \\
 & + \Delta t_{R3} + \Delta t_{13} + \Delta t_{R2} )
\end{split}
\end{equation}
\vspace{-0.08in}

As $Dancer \: 2$ is the $MSP$, from Fig.~\ref{fig:justificationSIA}, one can see $\Delta t_{12} + \Delta t_{23} = \Delta t_{13}$, and  $\Delta t_{1R} + \Delta t_{R3} = \Delta t_{13}$. Thus, Eq.~\ref{eq:justificationSIA} becomes:
\vspace{-0.08in}
\begin{equation} \label{eq:justificationSIA2}
min \: (\sum{\Delta t_{13} + \Delta t_{13} + \Delta t_{13} + \Delta t_{R2}} )
\vspace{-\topsep}
\vspace{1.5mm}
\end{equation}

As only the term $\Delta t_{R2}$ depends on the robot's movement in Equation~\ref{eq:justificationSIA2}, by minimizing $\Delta t_{R2}$ we can minimize the equation. 
Thus, if the robot and the $Dancer \: 2$ (in this case, the $MSP$) perform the same event at the same time, then $\Delta t_{R2}$ will become $0$, which will minimize Equation~\ref{eq:justificationSIA2}.
This implies that if the robot can perform the events close to the timings of the most synchronous person, then the whole group will reach a more synchronous state.

%%%%%%%%%%%%%%%%%%%%%%%%%%%%%%%%%%%%%%%%%%%%%%
%%%%%% Figure: Justification of SIA methods
%%%%%%%%%%%%%%%%%%%%%%%%%%%%%%%%%%%%%%%%%%%%%%

%%%%%%%%%%%%%%%%%%%%%%%%%%%%%%%%%%%%%%%%%%%%%%%%%%
%%%%%%%%%%%%%%%%%%%%%%%%%%%%%%%%%%%%%%%%%%%%%%%%%%

Thus, for a given iteration, $itr \in  \forall \: iterations$, the server will determine $MSP(itr_i)$. Then, during the next iteration, $itr_{(i+1)}$, the server will track all movements of $MSP(itr_{i})$. 
%and process those instructions to send to the robot in real-time.
The server then processes these information by utilizing the early detected events (\textit{early start moving forward}, \textit{early stop moving forward}, \textit{early start moving backward}, and \textit{early stop moving backward}) following the method described in the next paragraph.

As we know the timing of the events during the previous iteration of the dance $itr_{(i-1)}$, our anticipation method assumed that the similar event will happen more or less at the same time during this iteration $itr_{i}$.

Therefore, when it was close to the timing of events of $MSP(itr_{(i-1)})$ during $itr_{i}$, and the server received early detected events associated with $MSP(itr_{(i-1)})$, then the server anticipated those events as the indicator of the start of a movement.
The server then sent appropriate commands to the robot to perform that movement.

For example, suppose $Dancer \: 2$ was the most synchronous person during iteration 1, i.e., $MSP(itr_1)$ was $Dancer \: 2$.
$Dancer \: 2$ performed a \textit{start moving forward} event three seconds from the start of $itr_1$. 
So, during $itr_2$, it was assumed that the \textit{start moving forward} would happen three seconds from the iteration's start. 
Thus, if the server received a sufficient number of \textit{early start moving forward} events around $t_3$, then it notified the \textit{robot command generator} to generate commands to execute forward movement. 
This process was similar for all other regular events, excluding the clap event.

%%%%%%%%%%%%%%%%%%%%%%%%%%%%%%%%%%%%%%%%%%%%%%
%%%%%% Sub Section: 3.2
%%%%%% ECA Method
%%%%%%%%%%%%%%%%%%%%%%%%%%%%%%%%%%%%%%%%%%%%%%
\vspace{-0.08in}
\subsection{Event Cluster-Based Anticipation Method (ECA)}\label{sec:eca}
We created the ECA method to establish a reasonable comparison anticipation method for SIA that does not rely on group dynamics. ECA is theoretically simple, but powerful in nature. As the dance is rhythmic and iterative in nature, the movement events for one iteration are similar to events that happened in the previous iteration.

Thus, we averaged the events timing during one iteration to predict the timing of those same events for the next iteration.
Figure~\ref{fig:methods}-A explains this method visually.

First, for one iteration, we presented all the events associated with the dancers by a time series. 
Thus, this time series represented all the events of that iteration.
Then, we clustered all the similar types of events together those happened within a time threshold, $\epsilon$. 
For example, for a single event $e$, we calculated the timing of the event performed by three human participants, i.e., $t(dancer_1(itr_{i}),e)$, $t(dancer_2(itr_{i}),e)$, $t(dancer_3(itr_{i}),e)$.
Here, $t$ represents the timing of an event, and $itr_{i}$ represents the iteration $i$.

After that for each cluster of similar events, we calculated the average time of all the events and used that time as the timing of the event for the next iteration. 
These events and the times were the predicted events and timing for the next iteration of the dance. 
Thus, $t(robot(itr_{(i+1)}),e) = (t(dancer_1(itr_{i}),e) + t(dancer_2(itr_{i}),e) + t(dancer_3(itr_{i}),e)) / 3$. After the prediction of all the events for next iteration, the method sends a command to the \textit{robot command generator} module to generate an appropriate movement for the robot.

%%%%%%%%%%%%%%%%%%%%%%%%%%%%%%%%%%%%%%%%%%%%%%
%%%%%% Section: 4
%%%%%% Experiments
%%%%%%%%%%%%%%%%%%%%%%%%%%%%%%%%%%%%%%%%%%%%%%

\section{Experiments}\label{sec:experiment}

%%%%%%%%%%%%%%%%%%%%%%%%%%%%%%%%%%%%%%%%%%%%%%
%%%%%% Sub Section: 4.1
%%%%%% Pilot Studies
%%%%%%%%%%%%%%%%%%%%%%%%%%%%%%%%%%%%%%%%%%%%%%
\subsection{Pilot Studies}
Before performing the main experiment to compare performance between the two anticipation methods, we performed a series of pilot studies to test the setup of the system, and set various parameters for the two anticipation methods. We conducted two sets of pilot studies, with a total of seven participants (three women, four men). Participants were opportunistically recruited, and compensated with a \$5 gift card for participating \cite{Rack2015HRI}. 

During the first set of pilots, a sole participant danced with the robot. Here, we sought to measure two things: how fast the robot received action messages, and how accurately the robot performed with the human participant.

During the second set of pilot studies, a group of three participants danced with the robot. Here, we sought to establish appropriate parameters for the anticipation methods. To acquire these measurements, we recorded events generated from server logs as well as from odometry data from the robot. We compared the two, noting differences in velocity, distance, and event timings \cite{Rack2015HRI}.

%After the pilot study, we measured all the parameters necessary for the system.
%We also measured the optimal value of some parameters before running the main experiment.
Results from the pilot study showed that the robot received messages from the server within a timely manner. We also analyzed the movement patterns of the robot when it coordinated its movements with the humans, and found it to be well-coordinated.
Based on these data, we felt confident that the robot was moving synchronously with participants, and continued with the main experiment.

%%%%%%%%%%%%%%%%%%%%%%%%%%%%%%%%%%%%%%%%%%%%%%
%%%%%% Sub Section: 4.2
%%%%%% Main Experiment
%%%%%%%%%%%%%%%%%%%%%%%%%%%%%%%%%%%%%%%%%%%%%%
\vspace{-0.08in}
\subsection{Main experiment}
\label{main:exp}
We recruited a total of nine groups (27 participants in total, 3 persons per group) for our main experiment.
14 participants were women, 13 were men. Their average age was 22.93 years (s.d. = 3.98 years), and the majority were undergraduate and graduate students.
Only 3 participants had prior dancing experience, 24 did not.
Participants were recruited via mailing list and campus advertisement.
Upon scheduling a timeslot, participants were randomly assigned to join a group with two others. Each participant was compensated with an \$8 gift card for their time.

After giving informed consent, participants viewed an instructional video of the choreographed dance and the experimenters explained the different movements.
The participants then had time to practice the dance movements as a group as many times as they wanted. During this practice session, the robot did not dance with them.

Following the practice session, the group participated in three dance sessions. During the first session, only humans participated in the dance. During the last two sessions, the robot joined the group. In Sessions 2 and 3, the robot moved using either ECA then SIA, or SIA then ECA. (The order was counter-balanced to avoid bias). Participants were blind as to which method was in use.

During the last two sessions, the four clients recorded depth, infrared, and skeletal data of the participants, and the server logged all event and timing data. A single camera mounted on a tripod recorded standard video of the experiment for manual analysis purposes only.
%We did not record RGB data during the practice or the experimental sessions.

Following the experiment, participants completed a short questionnaire asking them to rate which of the two dance sessions they felt was more synchronous, a measure we have used in prior work \cite{Iqbal2015TAC}. Participants also reported which session they felt they were more synchronous with the rest of the group.
%We also asked them during which session they felt they were more synchronous with the rest of the group.

%%%%%%%%%%%%%%%%%%%%%%%%%%%%%%%%%%%%%%%%%%%%%%
%%%%%% Section: 5
%%%%%% Data Processing for Analysis
%%%%%%%%%%%%%%%%%%%%%%%%%%%%%%%%%%%%%%%%%%%%%%
%\section {Data Pre-Processing and Event Detection from the Robot}\label{sec:analysis}
\vspace{-0.05in}
\section {Robot Data Pre-Processing}\label{sec:analysis}

The server provided the human movement data logs and the clients raw data during the experiment, as detailed in Sections~\ref{sec:setup} and \ref{main:exp}. However, to conduct a complete comparison between the two anticipation methods, it is also necessary to determine how and when the robot actually moved during the two experimental sessions. To do this, we used the timestamped odometric data from the robot (\textit{x} and \textit{y} pose and $angular-z$ orientation), as well as the server-side robot command logs.

We calculated the same events for the robot as for the humans (forward, backward, stop, turn). 
Based on the changes in two consecutive \textit{x} or \textit{y} pose values and the robot's heading, we calculated whether the robot was moving forward or backward. 
For example, when the robot faced the first Kinect sensor and moved forward, then the changes in two consecutive pose values would be positive, if moving backward, negative. We detected turn events using changes greater than 0.4 in the \textit{z} heading value of the Turtlebot's \textit{angular twist} message. (Note, \textit{turn} events are considered equivalent to the humans' \textit{clap} events in our analysis.).

\textit{Stop} events were determined when a difference less than 0.002 was detected between two consecutive poses. These stop events were classified as forward or backward depending on the heading of the robot.

After detecting all events for the robot, we manually checked the data files for any errors.
During this process, we determined a 7\% missing event rate. 
%that approximately 7\% of events were missing in those files.
These missing events were evenly distributed across both of the anticipation methods. We manually checked the recorded video and odometric logs from the robot, and determined the robot actually moved correctly during the majority of those instances, so manually inserted the missing events into the files. There were a few instances (about 3.7\% overall) when the robot did not perform the activity that it was instructed to perform, which was mostly due to network latency. We discarded those data from the analysis.

%%%%%%%%%%%%%%%%%%%%%%%%%%%%%%%%%%%%%%%%%%%%%%
%%%%%% Section: 6
%%%%%% Data Analysis and Results
%%%%%%%%%%%%%%%%%%%%%%%%%%%%%%%%%%%%%%%%%%%%%%
\vspace{-0.05in}
\section {Data Analysis and Results}\label{sec:result}
To compare the performance and accuracy of the two anticipation methods, we first measured how synchronously the entire group, including the robot, coordinated their movements during both sessions. We then measured how appropriately timed the robot's motion was with its human counterparts.

%%%%%%%%%%%%%%%%%%%%%%%%%%%%%%%%%%%%%%%%%%%%%%
%%%%%% Sub Section: 6.2
%%%%%% Sync Index
%%%%%%%%%%%%%%%%%%%%%%%%%%%%%%%%%%%%%%%%%%%%%%
 
\vspace{-0.05in}
\subsection{Measuring Synchronization of the Group}
 Using the method described in \cite{Iqbal2015TAC} and discussed in Section~\ref{sec:sia}, we measured the degree of synchronization of the group for each iteration of the dance. 
First, we created individual time series for each of the dancers and the robot. Events in the time series were \textit{start moving forward}, \textit{stop moving forward}, \textit{start moving backward}, \textit{stop moving backward}, and \textit{clap}). 
Then, we calculated the pairwise synchronization index for each pair using the method described in Section~\ref{indiv}.

%%%%%%%%%%%%%%%%%%%%%%%%%%%%%%%%%%%%%%%%%%%%%%%%%%%%%%%%%%%%%%%%%%%%%%%%%%%%
%%%%%%%%%%%%%%%%%%%%%%%%%%%%%%%%%%%%%%%%%%%%%%%%%%%%%%%%%%%%%%%%%%%%%%%%%%%%
%%%%%% Sync Index Table
%%%%%%%%%%%%%%%%%%%%%%%%%%%%%%%%%%%%%%%%%%%%%%%%%%%%%%%%%%%%%%%%%%%%%%%%%%%%
%%%%%%%%%%%%%%%%%%%%%%%%%%%%%%%%%%%%%%%%%%%%%%%%%%%%%%%%%%%%%%%%%%%%%%%%%%%%

\begin{table}[t]
\centering
\caption{Group Synchronization Indices (GSI) for All Groups. Each group includes three people and one robot.}
\label{tab:sync_indx}
\begin{tabular}{CCCCCC}
\toprule
\multirow{2}{*}{\textbf{Group No.}} & \multicolumn{1}{c}{\multirow{2}{*}{\textbf{Iteration No.}}} & \multicolumn{2}{c}{\textbf{GSI}} & \multicolumn{2}{c}{\textbf{Mean GSI}} \\ \cmidrule(lr){3-4} \cmidrule(lr){5-6} 
 & \multicolumn{1}{c}{} & \multicolumn{1}{c}{\textbf{ECA}} & \multicolumn{1}{c}{\textbf{SIA}} & \multicolumn{1}{c}{\textbf{ECA}} & \multicolumn{1}{c}{\textbf{SIA}} \\ \midrule
\multirow{4}{*}{1} & 1 & 0.39 & {\bf 0.40} & \multirow{4}{*}{0.29} & \multirow{4}{*}{{\bf 0.45}} \\ 
 & 2 & {\bf 0.30} & 0.26 &  &  \\ 
 & 3 & 0.15 & {\bf 0.63} &  &  \\ 
 & 4 & 0.33 & {\bf 0.52} &  &  \\ \midrule
\multirow{4}{*}{2} & 1 & 0.28 & {\bf 0.66} & \multirow{4}{*}{0.46} & \multirow{4}{*}{{\bf 0.56}} \\ 
 & 2 & {\bf 0.39} & 0.37 &  &  \\ 
 & 3 & {\bf 0.63} & 0.51 &  &  \\ 
 & 4 & 0.54 & {\bf 0.71} &  &  \\ \midrule
\multirow{4}{*}{3} & 1 & 0.37 & {\bf 0.40} & \multirow{4}{*}{0.45} & \multirow{4}{*}{{\bf 0.52}} \\ 
 & 2 & 0.48 & {\bf 0.52} &  &  \\ 
 & 3 & 0.59 & {\bf 0.62} &  &  \\ 
 & 4 & 0.37 & {\bf 0.55} &  &  \\ \midrule
\multirow{4}{*}{4} & 1 & {\bf 0.30} & 0.20 & \multirow{4}{*}{{\bf 0.33}} & \multirow{4}{*}{0.29} \\ 
 & 2 & 0.37 & {\bf 0.38} &  &  \\ 
 & 3 & {\bf 0.37} & 0.30 &  &  \\ 
 & 4 & {\bf 0.28} & 0.28 &  &  \\ \midrule
\multirow{4}{*}{5} & 1 & {\bf 0.31} & 0.24 & \multirow{4}{*}{0.38} & \multirow{4}{*}{{\bf 0.43}} \\ 
 & 2 & 0.39 & {\bf 0.44} &  &  \\ 
 & 3 & 0.52 & {\bf 0.59} &  &  \\ 
 & 4 & 0.30 & {\bf 0.46} &  &  \\ \midrule
\multirow{4}{*}{6} & 1 & {\bf 0.50} & 0.43 & \multirow{4}{*}{0.41} & \multirow{4}{*}{{\bf 0.41}} \\ 
 & 2 & 0.41 & {\bf 0.45} &  &  \\
 & 3 & {\bf 0.40} & 0.32 &  &  \\
 & 4 & 0.33 & {\bf 0.46} &  &  \\ \midrule
\multirow{4}{*}{7} & 1 & {\bf 0.41} & 0.25 & \multirow{4}{*}{0.42} & \multirow{4}{*}{{\bf 0.47}} \\ 
 & 2 & {\bf 0.42} & 0.29 &  &  \\
 & 3 & 0.35 & {\bf 0.66} &  &  \\
 & 4 & 0.52 & {\bf 0.68} &  &  \\ \midrule
\multirow{4}{*}{8} & 1 & {\bf 0.78} & 0.32 & \multirow{4}{*}{0.50} & \multirow{4}{*}{{\bf 0.52}} \\ 
 & 2 & 0.40 & {\bf 0.48} &  &  \\ 
 & 3 & 0.41 & {\bf 0.72} &  &  \\ 
 & 4 & 0.41 & {\bf 0.56} &  &  \\ \midrule
\multirow{4}{*}{9} & 1 & {\bf 0.56} & 0.35 & \multirow{4}{*}{{\bf 0.45}} & \multirow{4}{*}{0.44} \\ 
 & 2 & {\bf 0.35} & 0.34 &  &  \\ 
 & 3 & 0.42 & {\bf 0.46} &  &  \\ 
 & 4 & 0.46 & {\bf 0.59} &  &  \\ \bottomrule
\end{tabular}
\vspace{-0.15in}
\end{table}
%%%
% END SYNC TABLE
%%%
 
From the pairwise synchronization index, we built a group topology graph ($GTG$) and calculated the individual synchronization index for each human dancer, as described in Section~\ref{sec:sia}. 
As the humans physically stood very close in proximity, we assumed that each of the group members was influenced by all other members of the group across the course of an entire dance session. (Every iteration, participants rotated their position, so a person in the front at $itr_i$ will end up in the back by $itr_{(i+2)}$.) Thus, in the analysis every human was connected in the graph with all other members of the group, including the robot.

When calculating the robot's individual synchronization index, we employed slightly different analyses between ECA and SIA. For ECA, because the robot's motion was based on the average of all dancers' motions in the previous iteration, when building the $GTG$ all edges from the robot connected to all other human group members. However, for SIA, at any given $itr_i$ the robot was only ever following $MSP(itr_{(i-1)})$ in real time. Thus, during $itr_i$ the robot was only influenced by that person, not by the other group members. Thus, it is logical to take only the pairwise synchronization index between the robot and that person into account while calculating the individual synchronization index of the robot and building the $GTG$ for that iteration.
Therefore, we only considered an outgoing edge from the robot to $MSP(itr_{(i-1)})$ in the $GTG$.
%Using the method described in Section~\ref{sec:sia}, we then calculated the individual synchronization index of the robot.

After measuring the individual synchronization index, we calculated the group synchronization index for each group using the method described in \cite{Iqbal2015TAC}.
Here, we describe the method very briefly.

While calculating the group synchronization index, both the individual synchronization index as well as the members' connectivity to the group  was taken into consideration. 
For a given vertex in the $GTG$, the ratio of the number of outgoing edges connecting to it, and the number of maximum possible edges in a very synchronized condition for that vertex, is called \textit{the connectivity value} ($CV$).
Thus we can define $CV$ of series $s_i$ as:

\begin{equation}
{CV(s_i)}  = \frac{\sum_{j=1,\ldots,H, \; j \neq i}{f(s_i,s_j)}}{H-1}
\end{equation}

The $CV$ represents how well an individual is synchronized with the rest of the group.
First, we calculate each individual's synchronization index multiplied by their $CV$. 
Then, the overall group synchronization index is computed by taking the average of this product \cite{Iqbal2015TAC}. 
So, the overall group synchronization index, $G_\tau$, is computed by: 

\begin{equation}\label{eq:overallSync}
G_\tau  = \frac{\sum_{i=1,\ldots,H}{I_\tau(s_i)} \times {CV(s_i)}}{H}
\end{equation}

While calculating the group synchronization index, we used $\tau = 0.25 s$. 
This value means we considered two events synchronous when the same types of events in two time series occurred within 0.25 seconds of one another.
 
%%%%%%%%%%%%%%%%%%%%%%%%%%%%%%%%%%%%%	
%%%%%% Graphs: Closeness to the Appropriate Timing 
%%%%%%%%%%%%%%%%%%%%%%%%%%%%%%%%%%%%%

%%%%%%%%%%%%%%%%%%%%%%%%%%%%%%%%%%%%%%%%%%%%%%
%%%%%% Figure: Timing Appropriateness
%%%%%%%%%%%%%%%%%%%%%%%%%%%%%%%%%%%%%%%%%%%%%%

\begin{figure}[b]
\vspace{-0.15in}
\centerline{\includegraphics[width=0.4\textwidth]{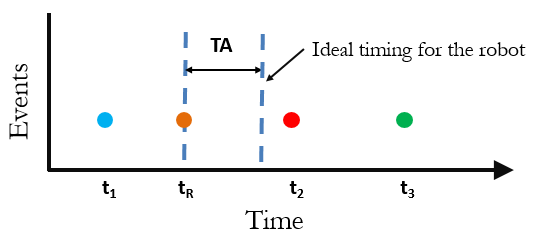}}
	\caption{Timing appropriateness calculation for the robot's movement.}
\label{fig:TA}
\vspace{-0.15in}
\end{figure}

Table~\ref{tab:sync_indx} presents the group synchronization indices (GSI) for each group (three humans and one robot), across both anticipation methods (ECA and SIA), and across the four iterations per session). The table also presents the average GSI for each group in the rightmost column. Boldface is used to indicate which of the two methods yielded a higher GSI, and this is indicated for both the per-iteration GSI and the average GSI per group.

For 22 out of 36 total individual dance iterations, the SIA method yielded a higher GSI than the ECA method. And in 7 out of 9 trials, the SIA method yielded a higher GSI than the ECA method.\footnote{Note, due to a small sample size (\textit{n} = 36), it would be dubious to run statistical means comparisons, and one should not accept a \textit{p-value} with certainty \cite{gelman2013commentary}. Instead, we agree with Gelman \cite{gelman2013commentary} that reliable patterns can be found by averaging, as reported here.}

Using a discrete analogue scale, we asked participants to rate on a scale from 1-5 how synchronous they thought the robot was with the other humans during the sessions. 
Based on their responses, we measured the more synchronous session of that trial, for which 2 out of 3 dancers agreed on their rating. 
For 7 / 9 trials, this collective rating matched with the more synchronous session of the trials determined by our method (See Table~\ref{tab:sync_indx}, last two columns.)
 
%%%%%%%%%%%%%%%%%%%%%%%%%%%%%%%%%%%%%%%%%%%%%%
%%%%%% Sub Section: 6.2
%%%%%% Closeness to the Appropriate Timing
%%%%%%%%%%%%%%%%%%%%%%%%%%%%%%%%%%%%%%%%%%%%%%
\vspace{-0.08in}
\subsection{Measuring Robot Timing Appropriateness}
For both anticipation methods, we aimed to have the robot perform its actions (events) as close as possible in time to its human counterparts.
Thus, we measured how close the robot's actual movement was to what the humans were doing at that time.

Thus, as a measure of timing appropriateness of the robot, we calculated the absolute time difference between the time when the robot performed an event, and the ideal timing of that event.
As a measure of the ideal timing of an event, we took the average timing of an event performed by the humans.
This measure is similar to the absolute offset measure used in \cite{Hoffman2010ROMAN}, however, the timing appropriateness measure used here is within the context of a group.

First, we represented all events associated with the humans during an iteration by a time series. 
Then, we clustered all the similar types of events together with those that were performed by the dancers within a time threshold, $\epsilon$. 
For example, for a single event $e$, we calculated the timing of the event performed by three human participants within $\epsilon$, i.e., $t(dancer_1,e)$, $t(dancer_2,e)$, $t(dancer_2,e)$.
We also calculated the timing of that event performed by the robot, $t(robot,e)$.
Then, to calculate the ideal timing for the robot, we take the average of these times of this event performed by the humans.
Thus, $t(robot_{ideal},e) = (t(dancer_1,e) + t(dancer_2,e) + t(dancer_3,e)) / 3$.
Then, we calculated the timing appropriateness ($TA$) of that event performed by the robot as, $TA(e) = \left|(t(robot,e) - t(robot_{ideal},e))\right|$. Figure~\ref{fig:TA} presents an example calculation of $TA$ for event $e$.

%%%%%%%%%%%%%%%%%%%%%%%%%%%%%%%%%%%%%%%%%%%%%%
%%%%%% Figure: Timing Appropriateness measures for CI
%%%%%%%%%%%%%%%%%%%%%%%%%%%%%%%%%%%%%%%%%%%%%%
\begin{figure}[b]
\vspace{-0.15in}
\centerline{\includegraphics[width=0.35\textwidth]{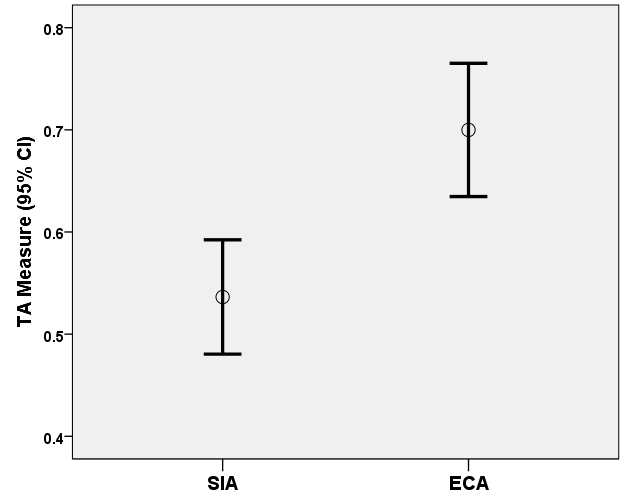}}
	\caption{Timing appropriateness measure for the robot with 95\% confidence interval for both methods, SIA and ECA.}
\label{fig:TAGraph}
\vspace{-0.15in}
\end{figure}

%%%%%%%%%%%%%%%%%%%%%%%%%%%%%%%%%%%%%%%%%%%%%%
%%%%%% Figure: Setup
%%%%%%%%%%%%%%%%%%%%%%%%%%%%%%%%%%%%%%%%%%%%%%
\begin{figure*}[t]
\centerline{\includegraphics[width=0.9\textwidth]{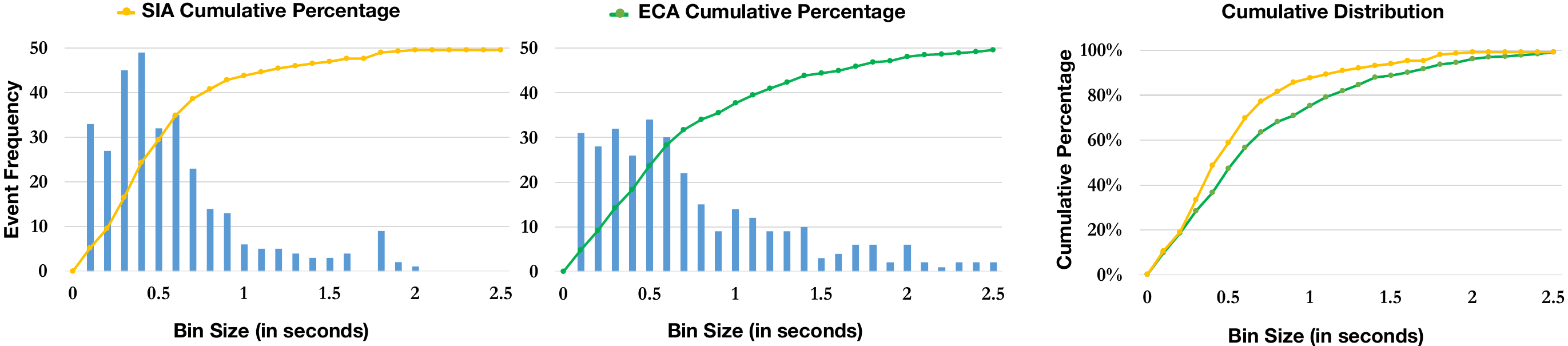}}
	\caption{Event frequency distribution and the cumulative percentage distribution of the timing appropriateness measure for the two anticipation methods. SIA (left) and ECA  (right). The rightmost graph shows the distribution of the timing appropriateness measure for both methods.}
\label{fig:hist}
\vspace{-0.15in}
\end{figure*}

%%%%%%%%%%%%%%%%%%%%%%%%%%%%%%%%%%%%%%
%%%%%% Results of the Sync Index Analysis
%%%%%%%%%%%%%%%%%%%%%%%%%%%%%%%%%%%%%%

After calculating $TA$ for each event during all the trials, we created two histograms, one for each anticipation method.
We used a bin size of 0.1 seconds, starting at $0 s$ and going to $2.5 s$. 
Then, we calculated the frequency of the events for which the $TA$ falls within that time span.

%%%%%%%%%%%%%%%%%%%%%%%%%%%%%%%%%%%%%%%%%%%
% Start tables
%%%%%%%%%%%%%%%%%%%%%%%%%%%%%%%%%%%%%%%%%%

%\begin{figure*}[tb]
%	\centering
%	\small
%	\subfigure{
%	\includegraphics[scale=0.2]{sia_hist4.png}
%	\label{fig:eca_hist}}
%	\hspace{-0.08in}
%	\subfigure{
%	\includegraphics[scale=0.2]{eca_hist4.png}
%	\label{fig:sia_hist}}
%	\hspace{-0.08in}
 %	\subfigure{
%	\includegraphics[scale=0.2]{cumulative4.png}
%	\label{fig:cmm_hist}}
% 	\hspace{-0.08in}
% 	\caption{Event frequency distribution and the cumulative percentage distribution of the timing appropriateness of events in seconds for the two anticipation methods (A) SIA method \& B) ECA method). Cumulative (\%) distribution of the timing appropriateness of events in seconds for both methods (C).}
%\label{fig:hist}
%\hspace{-0.15in}
%\end{figure*}

%%%%%%%%%%%%%%%%%%%%%%%%%%%%%%%%%%%%%%%%%%%
% End tables
%%%%%%%%%%%%%%%%%%%%%%%%%%%%%%%%%%%%%%%%%%

%%%%%%%%%%%%%%%%%%%%%%%%%%%%%%%%%%%%%	%%%%%% Results of the Closeness to the Appropriate Timing
%%%%%%%%%%%%%%%%%%%%%%%%%%%%%%%%%%%%%
In Figures~\ref{fig:hist}-A and B, we present histograms representing the timing appropriateness measure, and the cumulative percentage of event frequencies, for the ECA and SIA methods respectively.
Figure~\ref{fig:hist}-A (ECA), shows that the robot was able to perform 81.88\% of its events within $1.2 s$, and 90\% of its events within $1.6 s$ of the appropriate timing.
Figure~\ref{fig:hist}-B (SIA) shows that the robot performed 81.65\% of the events within $0.8 s$, and 90.82\% of the events within $1.2 s$ of the appropriate timing.

Figure~\ref{fig:hist}-C presents the cumulative percentage of events for both methods together.
One can find that the robot performed the events more appropriately during the SIA method, than compared to the ECA method.

%%\begin{figure}[t]
%%\centerline{\includegraphics[width=.45\textwidth]{ECA_Hist.png}}
%%\caption{Distribution and the cumulative (\%) distribution of the closeness to the appropriate timing of events in second for the ECA method}
%%\label{fig:eca}
%%\end{figure}

%%\begin{figure}[t]
%%\centerline{\includegraphics[width=.45\textwidth]{SIA_Hist.png}}
%%\caption{Distribution and the cumulative (\%) distribution of the closeness to the appropriate timing of events in second for the SIA method}
%%\label{fig:eca}
%%\end{figure}

%%\begin{figure}[t]
%%\centerline{\includegraphics[width=.45\textwidth]{cumulative.png}}
%%\caption{Cumulative (\%) distribution of the  closeness to the appropriate timing of events in second for both methods.}
%%\label{fig:eca}
%%\end{figure}

%Fig: A) Distribution and the cumulative (\%) distribution of the closeness to the appropriate timing of events in second for the ECA method, B) Distribution and the cumulative (\%) distribution of the closeness to the appropriate timing of events in second for the SIA method, C)  Cumulative (\%) distribution of the  closeness to the appropriate timing of events in second for both methods.

For the SIA method, the mean for the timing appropriateness measure was $0.54 s$ (s.d. = $0.59 s$) (See Figure~\ref{fig:TAGraph}). For the ECA method, the mean timing appropriateness measure was $0.70 s$ (s.d. = $0.50 s$) (See Figure~\ref{fig:TAGraph}). While these data did not have a normal distribution, as is visible from the graph and a normality test, they did have a sufficient number of means to compare statistically. We conducted a Wicoxon Signed Rank Test, and found that the timing appropriateness values for the ECA method were significantly larger than for the SIA method, $z=-4.399, p<0.05, r=-0.18$. This means that when using the SIA method, the robot moved more appropriately in time than when using the ECA method.

%%%%%%%%%%%%%%%%%%%%%%%%%%%%%%%%%%%%%%%%%%%%%%
%%%%%% Sub Section: 7
%%%%%% Discussion and Future work
%%%%%%%%%%%%%%%%%%%%%%%%%%%%%%%%%%%%%%%%%%%%%%
\vspace{-0.06in}
\section{Discussion and Future Work}\label{sec:discussion}

%%%%%%%%%%%%%%%%%%%%%%%%%%%%%%%%%%%%%%%%%%%%%%
%%%%%% Discussion
%%%%%%%%%%%%%%%%%%%%%%%%%%%%%%%%%%%%%%%%%%%%%%
The results suggest that the human-robot team was more synchronous using  SIA than using the ECA method. Moreover, when SIA was used, the robot was able to perform its actions significantly closer to the appropriate timing of the event. This supports the idea that SIA is well-suited to provide movement coordination information to a robot during an SJA scenario.

%%%%%%%%%%%%%%%%%%%%%%%%%%%%%%%%%%%%%	
%%%%%% Table: Closeness to the Appropriate Timing 
%%%%%%%%%%%%%%%%%%%%%%%%%%%%%%%%%%%%%
%\setlength{\tabcolsep}{13pt}
%\begin{table}[t]
%\caption{Mean and s.d. value of the Closeness to the Appropriate Timing for Both Methods}
%\label{tab:closeness}
%\centering
%\begin{tabular}{ccc}
%\midrule
%              & \textbf{Mean} & \textbf{s.d.} \\ %\midrule
%\textbf{ECA} & 0.70   & 0.59  \\
%\textbf{SIA}   & 0.54   & 0.50  \\
%\bottomrule
%\end{tabular}
%\end{table}

Additionally, these results might support the robustness of the SIA method over the ECA method, as the SIA method is more dynamic and adaptable to change within the group. In our study, the SIA method chose the most synchronous dancer in the group, and used that to inform the robot's actions in real-time. However, relying on a method like ECA would mean that if a dancer was moving asynchronously within the group, the robot's motion could be adversely affected (as it is following everyone). SIA is robust to handle this phenomenon, as a person who performed asynchronous movements within the group is unlikely to ever be chosen as the most synchronous person.

This work shows that taking team dynamics into account can be useful for robots when conducting coordinated activities with teammates. This work can lead others in the robotics community towards further investigating the role of a group on behavior, rather than just focusing on individuals. This has implications not only for human-robot interaction, but also for multi-robot systems research. 
We are currently exploring the effect of different anticipation methods in multi-human multi-robot scenarios \cite{Iqbal2015ICMI,Iqbal2016RSS}.

One limitation of this work is how event detection is calculated. In the current setup, a predefined set of human activities were detected by the system to understand the group dynamics. Building on this foundation, our future work will include incorporating human gross motion directly to the synchronization measurement step, instead of using pre-labelled events. 
Moreover, we are also planning to incorporate a decision module for robots, which will use the perceived knowledge to select the best decision from a set of options, based on the context \cite{aastha,maria-cbar}.

Another limitation of the current method is how it uses team metrics, and task-related information. For example, the method does not yet incorporate dancer expertise, nor does it factor in the tempo or dynamics of the music. In the future, we plan to incorporate an understanding of these factors. For example, in a team of novice dancers, a robot could perhaps keep a team on tempo.

%%%%%%%%%%%%%%%%%%%%%%%%%%%%%%%%%%%%%%%%%%%%%%
%%%%%% Future Work: These three para are same as the ICMI DC future work
%%%%%%%%%%%%%%%%%%%%%%%%%%%%%%%%%%%%%%%%%%%%%%

In the future we also seek to explore the use of robot-centric vision and local coordination methods to calculate synchrony. This will enable robots to operate in more dynamic settings, and lessen the need for external sensors.
However, incorporating local sensor data will be more challenging as it might be more noisy due to occlusion and local movements. However, we will build on our prior multimodal fusion and others' robot-centric perception work to overcome this challenge \cite{aastha,Ryoo2013}.

%Our future work will also include developing methods to detect high-level activities using the robot's on-board multimodal sensors. Incorporating both local and global sensor data will be helpful for better detection of high-level events, which will lead the robot to perceive  group dynamics more accurately. Incorporating local sensor data will be more challenging as it might be more noisy in nature due to occlusion and jerky movements. However, we will build on our prior multimodal fusion work as well as robot-centric perception work to overcome this challenge \cite{aastha,Ryoo2013}.

We also will explore incorporating other synchronization methods humans employ, such as adapting to continuous tempo changes, within the SIA algorithm. Models like ADAM (ADaptation and Anticipation Model) have been proposed in the literature to computationally model this behavior in humans by combining adaptation and anticipation during an activity \cite{VanDerSteen2013,VanDerSteen2015}.
It may be beneficial for a robot to have this ability both in human-robot and multi-robot teams.
%We also might explore the adaptation mechanism like those proposed in ADAM, but in the context of human-robot team.
%We might extend the error correction mechanism, and can integrate with our SIA algorithm. 
This integration might make the SIA algorithm more robust in anticipating, and synthesizing future activities more accurately. 

We also hope to extend our methods to work beyond SJA activities, such as timed but varied collaborative tasks within industrial settings. A human-robot team working in an industrial setting has specific sequences of activities to perform overtime, some of which might be independent, and might not have to happen synchronously. However, the events do have to happen contingently; so some of our anticipatory methods may be applicable.

Movement coordination is an important, emerging research area in robotics, neuroscience, biology, and many other fields \cite{gelblum2015ant,Bandouch2012,Zhang2011,Koppula2013,Ofli2013,Samadani2013}. Our work helps enable robots to have a better understanding of how to coordinate with the environment. This can be useful both for solving problems in robotics, and perhaps also in fields beyond.


%%%%%%%%%%%%%%%%%%%%%%%%%%%%%%%%%%%%%%%%%%%%%%%%%%%%%%%%%%%%%%%%%%%%%%%%%%%%%%%%

%%%%%%%%%%%%%%%%%%%%%%%%%%%%%%%%%%%%%%%%%%%%%%%%%%%%%%%%%%%%%%%%%%%%%%%%%%%%%%%%

%%%%%%%%%%%%%%%%%%%%%%%%%%%%%%%%%%%%%%%%%%%%%%%%%%%%%%%%%%%%%%%%%%%%%%%%%%%%%%%%

\section*{ACKNOWLEDGMENT}

The authors thank Afzal Hossain, Olivia Choudhury, James Delaney, Cory Hayes, and Michael Gonzales for their assistance.

%%%%%%%%%%%%%%%%%%%%%%%%%%%%%%%%%%%%%%%%%%%%%%%%%%%%%%%%%%%%%%%%%%%%%%%%%%%%%%%%
\balance

%\bibliographystyle{abbrv}
%\bibliography{mybib1}  % sigproc.bib is the name of the Bibliography in this case

%%%%%%%%%%%%%%%%%%%%%%%%%%%%%%%%%%%%%%%%%
%BIB
%%%%%%%%%%%%%%%%%%%%%%%%%%%%%%%%%%%%%%%%
\bibliographystyle{IEEEtran}
\bibliography{IEEEabrv,mybib1}

%%%%%%%%%%%%%%%%%%%%%%%%%%%%%%%%%%%%%%%%
\begin{IEEEbiography}[{\includegraphics[width=1in,height=1.25in,clip,keepaspectratio]{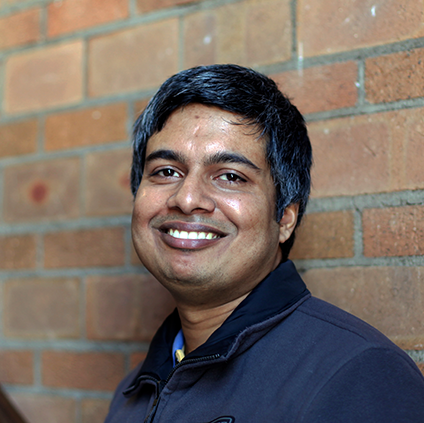}}]%
%\begin{IEEEbiographynophoto}
{Tariq Iqbal} received the BS degree in computer science and engineering from Bangladesh University of Engineering and Technology, and the MS degree in computer science from the University of Texas at El Paso. He is currently working toward the PhD degree in the Department of Computer Science and Engineering, University of Notre Dame. His research interests include robotics, computer vision, machine learning, and social signal processing. He is a student member of the IEEE.
\end{IEEEbiography}

%\vspace{-150 mm}
\vspace{-10 mm}

\begin{IEEEbiography}[{\includegraphics[width=1in,height=1.25in,clip,keepaspectratio]{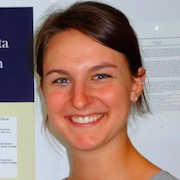}}]%
%\begin{IEEEbiographynophoto}
{Samantha Rack} is currently working toward the BS degree in computer science at the University of Notre Dame. She will begin working in industry following graduation. Her research interests include human-computer interaction and social signal processing. She is a student member of the IEEE.
\end{IEEEbiography}

\vspace{-10 mm}

\begin{IEEEbiography}[{\includegraphics[width=1in,height=1.25in,clip,keepaspectratio]{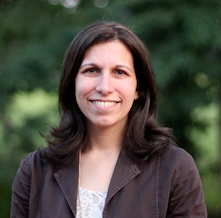}}]%
%\begin{IEEEbiographynophoto}
{Laurel D. Riek} (M'00-SM'14) received the BS degree in Logic and Computation from Carnegie Mellon University and the PhD degree in Computer Science from the University of Cambridge. She is the Clare Boothe Luce Assistant Professor of Computer Science and Engineering at the University of Notre Dame, and directs of the Robotics, Health, and Communication Laboratory. Her current research interests include robotics, social signal processing, human behavior understanding, and healthcare informatics. She is the author of more than 75 technical articles in these areas. She serves on the editorial board of IEEE Transactions on Human Machine Systems, the Steering Committee of the ACM/IEEE Conference on Human-Robot Interaction, and numerous conference program committees. She is a senior member of the IEEE. 
\end{IEEEbiography}

\end{document}